\documentclass{article}
\usepackage{ijcai24}

\usepackage{times}
\usepackage{soul}
\usepackage{url}
\usepackage[hidelinks]{hyperref}
\usepackage[utf8]{inputenc} 
\usepackage[small]{caption}
\usepackage{graphicx}
\usepackage{amsmath}
\usepackage{amsthm}
\usepackage{booktabs}
\usepackage{algorithm}
\usepackage{algorithmic}
\usepackage[switch]{lineno}

\usepackage{bbm}
\usepackage{bm}
\usepackage{amsfonts}
\usepackage{float}
\usepackage{wrapfig}

\DeclareMathOperator*{\argmin}{arg\,min}

\title{Towards Integrating Personal Knowledge into Test-Time Predictions}

\author{
Isaac Lage$^1$\and
Sonali Parbhoo$^2$\And
Finale Doshi-Velez$^3$\\
\affiliations
$^1$Colby College\\
$^2$Imperial College London\\
$^3$Harvard University\\
\emails 
ilage@colby.edu,
s.parbhoo@imperial.ac.uk,
finale@seas.harvard.edu
}

\begin{document}

\maketitle

\begin{abstract}
Machine learning (ML) models can make decisions based on large amounts of data, but they can be missing personal knowledge available to human users about whom predictions are made. For example, a model trained to predict psychiatric outcomes may know nothing about a patient's social support system, and social support may look different for different patients. In this work, we introduce the problem of \textit{human feature integration}, which provides a way to incorporate important personal-knowledge from users without domain expertise into ML predictions. We characterize this problem through illustrative user stories and comparisons to existing approaches; we formally describe this problem in a way that paves the ground for future technical solutions; and we provide a proof-of-concept study of a simple version of a solution to this problem in a semi-realistic setting. 
\end{abstract}

\section{Introduction}
\label{sec:introduction}

ML models are attractive partly because they promise personalized predictions and decisions at scale: whether it is credit score or a recipe recommendation, outputs can be provided to users quickly and with accuracy. These outputs are typically produced based on a pre-specified set of inputs. Indeed, ML models are often regularized to make generally accurate predictions across a population with a minimal number of inputs, and some data regulations such as the GDPR require collection of only minimal inputs \cite{GDPR2016a}. However, the inputs that are predictive on average may not tell the full story for any individual. In fact, to accurately represent a user's situation, we may need access to features not in the dataset at all.

For example, consider the question of trying to predict depression-related outcomes for mental health patients. A typical approach (e.g. in \cite{pradier2020dropout} or \cite{lage2022efficiently}) may involve training an ML model given the electronic health record (EHR) data from thousands of patients. These EHRs capture hundreds of variables related to the patient's diagnoses, treatments, and utilization patterns. However, patients living with depression have access to \textit{personal knowledge} that may be important: one patient may have a job that interferes with their sleep; another may have a poor support network; another may have limited access to transportation \cite{carter2012comprehensive}. Building ML models that can account for this context could lead to more accurate predictions and better user experiences.

\textit{Personal knowledge}, like in this example, generally consists of lived experience, which can be deeply important for predicting events in a person's life \cite{suresh2021beyond}. This is in contrast to \textit{formal knowledge}--related to codified theories, and \textit{instrumental knowledge}--related to the application of theories. In this work, we propose a novel problem formulation--\textit{human feature integration}, designed to provide pathways for users to incorporate personal knowledge in the form of new features not otherwise in the dataset into predictions made \textit{about} them. Human feature integration does not require the user to have formal or instrumental knowledge about how to make a prediction, only \textit{personal knowledge} about their lived experience.

The two key aspects of the human feature integration problem are that there exist \emph{human features} that are not used by the ML system in its predictions that consist of (a) information available to the person \emph{about whom} a prediction is made (e.g. the patient, rather than the doctor or ML expert) and (b) whose relevance varies across individuals (e.g. for patient A their sleep matters, but for patient B their support network does). These 2 aspects require novel methods, as (a) means the user is generally \textit{not} an expert decision maker, so methods that rely on human \textit{predictions} (e.g. \cite{raghu2019algorithmic}) are not applicable; and (b) means that we cannot simply add a small number of additional features to our data collection process as different features are required for different users (e.g. adding sleep and support network to a patient questionnaire is insufficient as patient C may require features related to their transportation).

In this work, we characterize the problem of human feature integration with the goal of encouraging future efforts towards addressing it. We believe these efforts will broaden the tools available to end-users of ML systems to retain agency over predictions made \textit{about} them. Towards this goal, we first characterize the importance of this problem through illustrative user stories and comparisons to existing approaches. We formally describe this problem in a way that paves the ground for future technical solutions. Finally, we provide a proof-of-concept study of a simple version of a solution to this problem in a semi-realistic setting. This study provides preliminary evidence towards the utility of solving this problem formulation, and a template for evaluating solutions to this problem.

\section{User Stories about Personal Knowledge}
\label{sec:user-stories}

We characterize this problem concretely through two fictional, illustrative user stories.  These describe scenarios where human features pertaining to personal knowledge are important when interacting with an ML system. We describe avenues provided to the user under existing approaches, and how human feature integration expands their options.

\paragraph{Scenario 1: Medical prediction for a transgender patient} A patient who is a transgender man taking testosterone came into the ER and was flagged as at risk of a future cardiac event by an ML model. The model's prediction is heavily influenced by the patient's hemoglobin levels being high for someone female, which is the gender recorded in the patient's medical record \cite{deutsch2013electronic}. As the model uses the lab values and demographic data collected in the ER and not the patient's full medical record from another healthcare system, it does not know about his testosterone treatment, and he has not disclosed it to the doctors in the ER. 

The patient does not have the medical expertise to estimate his risk of a future cardiac event, and also does not know that testosterone treatment can raise hemoglobin out of the expected range for patients with gender marker `f' \cite{velho2017effects}. So even if he saw an explanation of the prediction, he would not know that the prediction was being influenced by his testosterone treatment, and his ER doctor is unaware of the treatment as the patient does not know it is relevant.

Under the setting we propose, this patient could be queried by the model for whether he is on testosterone treatment based on his high levels of hemoglobin. He could input that he is and the model would update its prediction using that relevant personal expertise to give the patient a more accurate prediction of his risk of a future cardiac event.

\paragraph{Scenario 2: CPS flag for a child with aggressive outbursts} A family whose child has Disruptive Mood Dysregulation Disorder (DMDD) has been flagged for follow-up by Child Protective Services (CPS)--largely due to aggressive outbursts (loosely inspired by a participant quote in \cite{brown2019toward}). The child would benefit from specific psychiatric care for their DMDD rather than intervention by CPS, however the family does not have access to this care, and the model does not know that the child's outbursts are due to DMDD \cite{benarous2017evidence}. An explanation of the model's prediction would tell the family that the prediction depends mostly on the outbursts, but without domain knowledge about government agencies, they family would not know that having DMDD would route the child to psychiatric care rather than CPS. There is also the issue of whether the family would be believed when challenging a prediction due to the strong incentive to keep their child at home. 

Under the setting we propose, the ML model would query the family about whether their child has DMDD that causes the outbursts.  In this case, it would predict that the child should be treated for DMDD rather than subjecting the family to scrutiny by CPS. The family could provide documentation of the child's DMDD that could be verified as fact rather than questioned because of the complex incentives of the CPS system. This process would allow the family's personal knowledge about their child's medical history to influence the prediction and increase its accuracy without requiring the family to understand how decisions about government services are made, or have their incentives questioned.

\section{Alternative Approaches to Personal Knowledge (Require Domain Knowledge)}

In this section, we describe existing alternative approaches to combining personal knowledge with predictions from machine learning models, and demonstrate on a pedagogical toy example a case where they are insufficient. We then show on that example how our proposed problem formulation improves performance. This demonstrates that our problem formulation is technically novel and fills a gap in the literature. 

\subsection{Alternative Approaches: Related Work}

Existing methods to combine human and ML knowledge at the instance-level include approaches that combine human and ML predictions like mixture of experts, and explainable ML. Importantly, both of these require domain expertise.

\paragraph{Post-Hoc Prediction Combinations}
Methods exist for combining the predictions of multiple predictors to improve performance. Relevant in this setting are mixtures of experts when one expert is a human (e.g. \cite{pradier2021preferential}, \cite{parbhoo2017combining}), and methods that triage predictions between human and machine decision-makers (e.g. \cite{madras2018predict}, \cite{raghu2019algorithmic}). While these are quite effective in some settings, they can fail in cases where no individual decision-maker makes a correct prediction. For this method to work well, the user must often make accurate predictions--implying significant domain expertise.

\paragraph{Machine Learning Explanations} Interpretable ML models that provide explanations of their prediction process (e.g. \cite{wang2017bayesian}, \cite{lakkaraju2016interpretable}) allow an end-user to combine their knowledge with a prediction from an ML model. However doing this requires the user to understand how predictions \textit{should} be made in the problem domain, implying expertise at the prediction task. For example, the patient in user story 1 would need to understand both how the ML model is making its prediction, and how their feature--their testosterone treatment, \textit{should} influence the prediction in order to have enough information to accurately update the model's prediction. This is a hard problem, and requires a deep understanding of the domain.

\subsection{Our Problem Formulation Leads to Different Solutions: A Pedagogical Example}

Through a pedagogical example, we demonstrate that our problem formulation can produce novel behaviors in practice that are unachievable in some settings by these other 2 classes of approaches. This demonstrates that our problem formulation is technically novel, and provides intuition for how it differs from the two alternative approaches considered.

\begin{figure}
 \centering
 \includegraphics[width=0.33\textwidth]{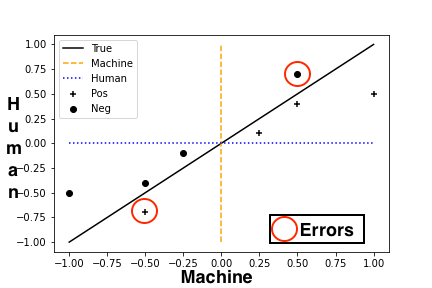}
 \caption{This figure shows a toy dataset with a human feature (y axis) and a machine feature (x-axis) that are both needed to produce the true decision boundary (solid line). The orange dashed line ($x=0$) and the blue dotted line ($y=0$) are the best fit lines based on the human and machine feature respectively. Both mis-classify the 2 red-circled points, but a model that is able to use both features can classify perfectly along $y=x$.}
 \label{fig:toy}
\end{figure}

Our toy example is a binary classifier with 2 features--one belonging to the machine feature set, and the other to the human feature set. Figure~\ref{fig:toy} shows this example, where the true labels depends on both features, and the best-fit lines for either feature alone mis-classify the same 2 points (circled).

\paragraph{Comparison to Post-Hoc Prediction Combinations} The 2 errors cannot be corrected by any weighted average of the human and machine-feature predictions because both make the same errors on these 2 misclassified points. If the weights can be negative, any output can be produced from 2 inputs, but importantly, this would mean that at least one of the predictions was being flipped--i.e. we are assuming that we know at least one of the decision-makers is wrong. In a mixture of experts context, we are assuming that both decision-makers are often accurate so allowing negative weights would not make sense. This demonstrates that these errors cannot be corrected with a post-hoc combination of the human and machine predictions.

\paragraph{Comparison to ML Explanations} The prediction based on the machine feature is interpretable (i.e. predict $1 \texttt{ if } x>0$), however this explanation cannot be easily combined with the human feature to correct the 2 misclassifications. A straightforward combination of the explained decision boundary and the human's decision boundary by taking the ``or'' of these decision rules is: predict $1 \texttt{ if } ( x>0 \texttt{ or } y>0)$. This does not address the errors, and the same is true if we combine the human and machine decision rules with ``and.'' Instead, fixing the erroneous predictions requires generating a completely new decision boundary using both the human and machine features. This is does not make use of the explanation of the machine learning model, and as such, relies entirely on the user's domain expertise. 

\paragraph{Our Problem Formulation: Human Feature Integration} If both features were available to the model when making the prediction, a linear classifier can easily separate the points perfectly. The key difference between this scenario and the previous one is that both features are provided to the \textit{machine learning system} which produces the prediction, while in the previous scenario, the 2 features were available to the \textit{user} who produced the prediction. Because in human feature integration, the user must only provide the relevant feature to the ML model, they are able to use their personal knowledge without being forced to generate a prediction themselves, which would require domain knowledge. These results demonstrate that feature-level integration can provide correct predictions in some cases where post-hoc prediction combinations and ML explanations cannot, showing that it is a distinct technical problem that fills a gap in the literature.

\section{Problem Definition-Eliciting Human Features at Test Time}
\label{sec:problem-def}

In this section, we propose a formalization of the problem of human feature integration. Our core technical intuition is that this problem can be formalized as the problem of how to query a small, but predictive set of human features for each instance, given that we have some way of making predictions based on these features. We first re-iterate problem requirements, then formalize those requirements as technical assumptions, and propose an idealized optimization objective to solve this problem. Finally, we describe 3 key technical challenges presented by our objective. 

\subsection{Problem Requirements}

In the previous sections, we explored the importance of personal knowledge on the part of the person \textit{about whom} a prediction is made. We formalize personal knowledge as human features--i.e. features where the value must be provided by the user of the model. As these must be queried from the individual user, then incorporated into that prediction, these features must be queried at test time. As such, we assume that after training, the ML model is fixed, and at test time, we can incorporate additional feature values queried from the user into our predictions.

The constraint of eliciting feature values from users at test time limits the amount of features that can be considered. While more features provide more information, too many queries to the user for feature values may overwhelm the user or interfere with the workflow in which the model is used--e.g. asking a patient 100 questions during a clinical interview may negatively impact the clinical relationship. 

Finally, we assume that the features that are important for each user will vary. Personal knowledge is highly contextual, and as such, the features that are informative for some users will be irrelevant for others. For example, the gender marker in the medical record of the patient in User Story 1 was important for him, but gender markers will not provide additional information for most patients who are cisgender. 

\textbf{We assume the existence and \textit{importance} of a small set of heterogeneous, instance-level human-features that we aim to elicit from users and incorporate into ML predictions at test time.} These features can represent personal knowledge that should influence the ML prediction. For example, these features can include a transgender patient's knowledge that they are on hormone therapy, or a family's knowledge that their child has been diagnosed with DMDD as we discuss in Section~\ref{sec:user-stories}.

\subsection{Technical Assumptions}

Formalizing the problem requirements described above leads us to the following set of technical assumptions. 

Concretely, we assume a standard supervised learning setup where we have a dataset of $N$ instances, the machine features, $X^m$, of dimensionality $D^m$, that the machine learning model always has access to, and corresponding labels, $Y$. 

We additionally assume the existence of a set of human features $X^h$, of dimensionality $D^h$, that must be queried at a cost during test time. We assume the user has access to these features for a specific instance about which a prediction is being made, as is the case when the instance corresponds to a representation of the user--for example their medical history. 

We assume that, for each instance, we can query the associated user for the values of specific features for the instance a small, fixed number of times $B$, for e.g. less than 10 times. 

We assume access at test time to a trained model that has parameters corresponding to $X^h$. I.e. we take as a given during test time the existence of a discriminative prediction function $f$ that makes predictions $\hat{Y}$ as follows:
\begin{equation}
 \hat{Y} = f(\theta^m, X^m, \theta^h, X^h) + \phi
\label{eqn:model}
\end{equation}
where $\theta^m$ and $\theta^h$ are parameters for $X^m$ and $X^h$ respectively and $\phi$ is a bias term. 

While the objective we propose in Section~\ref{sec:objective} takes $f$ as a given, how to fit $f$ during training remains a question. The $\theta^m$ term can be fit with a standard ML approach since $X^m$ is available at both train and test time. Fitting $\theta^h$ is a challenge since we assume that $X^h$ is expensive to obtain. As such, we consider fitting the $\theta^h$ term in $f$ as the first technical challenge involved in solving the objective described below, and consider strategies for doing so in Section~\ref{sec:challenges}. 

\subsection{Idealized Optimization Objective}
\label{sec:objective}

We formalize the problem of human feature integration as a constrained optimization problem with the goal to intelligently select $B$ human features to query at test-time, specific to each instance, that maximize the predictive quality, $L$, for that instance. 

For a test instance $x$, we denote which features in $x^h$ have been queried, using a binary query mask, $q \in \{0, 1\}^{D_h}$. $q_d = 1$ indicates that human feature $d$ has been queried for this instance, while $q_d = 0$ indicates that it has not been queried.

We can then write the constraint on the number of queries for each instance in terms of $q$:
\begin{equation}
\sum_{d=1}^{D_h} q_d \leq B
\end{equation}
In other words, the number of elements in $q$ that are queried, i.e. set to 1, should be at most $B$. 

To make predictions on a test instance where we only have access to queried features in $q$, we can marginalize out the remaining features (i.e. $q_d = 0$). This is a standard way of handling missing data. We denote this prediction function as $f^{marg}$, and use it to predict for an instance as follows:
\begin{equation}
\begin{aligned}{L}
 f^{marg}(x^m, x^h, q) = \\
 \int_{x^h_{q=0}} f(\theta^m, x^m, \theta^h, x^h_{q=1},x^h_{q=0}) p(x^h_{q=0}|x^m, x^h_{q=1})
\label{eqn:fmarg}
\end{aligned}
\end{equation}
Fitting an approximation to $p(x^h_{q=0}|x^m, x^h_{q=1})$ to compute this prediction function is the second key challenge as it also requires access to at least some human feature values. 

Computing the integral in Equation~\ref{eqn:fmarg} can be done using a standard approach of Monte Carlo sampling:
\begin{equation}
\begin{aligned}
 f^{marg}(x^m, x^h, q) \approx \frac{1}{S} \sum_{s=1}^S f(\theta^m, x^m, \theta^h, x^{h(s)}) \\ 
 x^{h(s)}_{q=1} = x^h_{q=1} \\
 x^{h(s)}_{q=0} \sim p(x^h_{q=0}|x^m, x^h_{q=1})
\end{aligned}
\end{equation}
where $S$ is the number of samples drawn. This corresponds to sampling the un-queried dimensions of $x^h$ and using the true values of the queried dimensions in each sample.

We can now define an idealized optimization objective that optimizes $q$ for instance $x$ to minimize predictive loss $L$ between the true label $y$ and the prediction made using only the machine features and the queried human features in $q$. This objective can be written as follow:
\begin{equation}
\begin{aligned}
 q* = \argmin_{\substack{%
 q \in \{0, 1\}^{D_h} \texttt{s.t.} \sum_{d=1}^{D_h} q_d \leq B}} L(y, f^{marg}(x^m, x^h, q))
\label{eqn:loss}
\end{aligned}
\end{equation}

While this is the goal we propose for human feature integration, we, by definition, do not have access to $y$ for any test instance $x$ to use in computing this objective. Effectively approximating this objective is the third key technical challenge.

\subsection{Key Technical Challenges}
\label{sec:challenges}

There are several key technical challenges that must be solved in order to operationalize a solution to the problem we described in Section~\ref{sec:objective}. 1) How can we get model parameters for the human features? 2) How can we approximate the probability of un-queried human features used in Equation~\ref{eqn:fmarg}? 3) How should approximate the idealized objective in Equation~\ref{eqn:loss} without knowing the label for test instances? We describe each of these challenges below.

\paragraph{Challenge 1: Model parameters for human features} By assumption, we assume that we have access to a predictive function $f$, shown in Equation~\ref{eqn:model}, that can be used to make predictions at test time given the machine features $X^m$ and the human features $X^h$. Said differently, we need model parameters corresponding to both $X^m$ and $X^h$. However, fitting this function is challenging as $X^h$ must be queried at a cost. This challenge can be approached in multiple ways including allocating a higher budget to querying human features at train time, or active-learning based methods for fitting parameters for $X^h$ with a few informative queries to $X^h$.

\paragraph{Challenge 2: Approximating the probability of un-queried human features} Computing $f^{marg}(x^m, x^h, q)$ requires approximating $p(x^h_{q=0}|x^m, x^h_{q=1})$--i.e. the probability of un-queried human features given the machine features and any queried human features. As in Challenge 1, fitting a model to approximate this distribution requires access to at least some of $X^h$, so similar methods can be employed. 

\paragraph{Challenge 3: Approximating the idealized objective without knowing y} The last key challenge is that Equation~\ref{eqn:loss} relies on labels for test instances. As this optimization happens at the level of each individual instance, it is not straightforward to use a training set to avoid needing the label at test time. One plausible approach to this challenge is active-learning based approaches that use uncertainty reduction rather than measuring the difference between the prediction and the true label (e.g. \cite{quost2021decision}. Another plausible approach could train a function to optimize the query set based on $X^m$ similar to \cite{jethani2021have}.

\section{Operationalizing an Approach to Human Feature Integration}
\label{sec:approach}

In Section~\ref{sec:problem-def}, we provide a general approach to the problem of using personal knowledge in ML predictions by querying informative, individualized sets of human features at test time. Here we describe one concrete but simple solution to that problem. While there are many opportunities to improve on the design choices we made, this demonstrates that solving this problem is feasible, and allows us to include preliminary experimental results demonstrating its utility on semi-realistic data.

\paragraph{Approach to Challenge 1: Assume access to $X^h$ During Training}

To address the challenge of acquiring model parameters for $X^h$, we assumed access to $X^h$ during training. This may be reasonable in settings where there is a budget for collecting additional human features at train time, including through crowdsourcing (e.g. \cite{cheng2015flock}), or in clinical trials prior to integration into clinical practice (e.g. the CoMMpass study for multiple myeloma \url{https://themmrf.org/finding-a-cure/our-work/the-mmrf-commpass-study/}
).

\paragraph{Approach to Challenge 2: Assume access to $X^h$ During Training and Independence}

We address the challenge of approximating the probability distribution $p(x^h_{q=0}|x^m, x^h_{q=1})$ required to compute Equation~\ref{eqn:fmarg} by again assuming access to $X^h$ at train time, as in Challenge 1. We also make a computational simplification by modeling $p(x^h_{q=0}|x^m)$ instead of $p(x^h_{q=0}|x^m, x^h_{q=1})$--i.e. removing the conditioning on the queried human features. This allows us to fit an independent model to prediction each dimension of $X^h$ given $X^m$. While this assumption--that the dimensions of the human features are independent, is likely not true in practice, we assume so as not to have to either marginalize out the unqueried human features during training, or train models for each subset of human features. The results from our preliminary experiments would likely be improved if the correlations between $X^h$ had been modeled, but even so, our results show improvements over alternative approaches to using personal knowledge.

\paragraph{Approach to Challenge 3: Minimize Predictive Entropy}

In the absence of true labels, we take the approach of minimizing the expected entropy of the marginalized predictions instead of the difference between the predicted labels and true labels as shown in Equation~\ref{eqn:loss}. This corresponds to finding the human features to query for each instance that reduce the prediction uncertainty as much as possible. Taking the expectation of the entropy w.r.t the unqueried features approximates their utility before paying the cost to query a feature.

We use the method introduced by \cite{quost2021decision}, that iteratively selects the query $d^*$ for an instance that minimizes the expected entropy of the marginalized predictions. Formally, the approach solves the following optimization problem for each feature query for each instance:
\begin{equation}
d^* = \argmin_{d \in \{1,...,D^h|q_d=0\}} \mathbb{E}_{p(x^h_d|x^m)}[H(f^{marg}(x^m, x^h, q + \mathbbm{1}^{D^h}_d)]
\label{eqn:subproblem}
\end{equation}
where $\mathbbm{1}^{D^h}_d$ denotes a onehot vector of size $D^h$ and all entries are $0$ except at index $d$ where the entry is $1$. 

In order to simplify the computation of the expectation, we we assume $X^h$ is binary: $X^h \in \{0, 1\}^{N\texttt{x}D_h}$. We additionally assume that the labels are categorical, i.e. $Y_n \in \{0, ..., K\}$. This allows us to compute the expectation as a summation over these discrete values.

To optimize the full query set defined by $q$, the approach greedily repeats this procedure for each query in the budget $B$. While this is not guaranteed to provide an optimal solution, it is often employed for similar approaches, e.g. \cite{houlsby2011bayesian}, and is generally considered effective. 

\section{Preliminary Experiments}

In this section, we use the simple solution discussed in Section~\ref{sec:approach} to conduct preliminary experiments towards better understanding the practical implications of the problem we study. We first lay out three research questions designed to help us understand if, when, and how solving our proposed problem can improve over baselines. Then, we describe key aspects of our experimental setup. Finally, we describe our results, organized by research question. These experiments provide two things. 1) Initial evidence towards the utility of human feature integration. 2) A framework that can be used to evaluate approaches to human feature integration.

\subsection{Research Questions}

We lay out 3 research questions that form the basis of our experimental evaluation.

\paragraph{RQ1: Can a few human features improve predictions?} Can incorporating human features improve predictions over those made with only machine features? If so, can this improvement be obtained with only a small ($\leq B$) subset of the human features? If additional features hold no utility, then methods to incorporate them into ML predictions will not improve performance. Similarly, if they are only useful in large numbers, human feature integration will not improve performance.

\paragraph{RQ2: Does instance-level feature selection provide value over dataset-level feature selection?} Do the features needed differ across instances, or does the same set of human features work well enough for all instances? If the latter, human feature integration is unnecessary and updating the dataset to include these important features may yield better results.

\paragraph{RQ3: Is choosing the right set of human features better handled by users or algorithms?} If users can easily tell which subset of their features will improve predictions, then developing algorithms to select those features may not be necessary. But if selecting the right human features is challenging for users, then querying users with an algorithm, as in our proposed approach, may be a better.

\subsection{Experimental Setup}

To answer these research questions experimentally, we use the approach described in Section~\ref{sec:approach} and compare it to baselines. We apply it in real datasets with synthetically generated splits between human and machine features. I.e. the human features are real features, but they didn't come from querying a human during test time, and we have access to them when evaluating our experiments. See Appendix for more detail.

\paragraph{Datasets} We use two real datasets in our experiments: a \textit{recipe} dataset
\footnote{\url{https://www.kaggle.com/datasets/kaggle/recipe-ingredients-dataset} train.json file} where the goal is to predict the cuisine of a recipe (e.g. Italian food) based on the ingredients (6k instances, 120 features, and 20 classes after pre-processing), and a \textit{birds} dataset (Wah et al. 2011) 
 where the goal is to predict the type of bird (e.g. crow) based on some crowdsourced attributes of an image of a bird (5k instances, 171 features, and 36 classes after pre-prerocessing). 

Rather than running user studies to collect the human features, we synthetically create sets of human and machine features by assigning some of the features in both datasets to the human features--$X^h$, and some to the machine features--$X^m$. In both cases, we constructed these feature splits so that $X^m$ consists of a smaller set of features that are ``simpler'' in some way, and $X^h$ consists of a larger set of features that are more ``complex'', better simulating contextual, personal knowledge. In the \textit{recipe} dataset, we split the features so all single word ingredients (33 features) are in the machine feature set, and all 2$+$ word ingredients (87 features) are in the human feature set. For the \textit{birds} dataset, we split them so all non-color-related words (48 features) are in the machine features, and all color-related words (123 features) are in the human features.

\paragraph{Proposed Approach}

We implement the approach in Section~\ref{sec:approach} with logistic regression models for $f$ and the individual dimensions of $\hat{p}(X^h_d|X^m)$. We call this method \textit{entropy-selection}. We also include a variant of this approach with a re-training heuristic described in the Appendix called \textit{entropy-retrain}. The results suggesting the need for the retraining heuristic are also in the Appendix.

\paragraph{Baselines}

We compare the proposed solutions to four baselines and one oracle upper bound. For all of these methods, we use logistic regression models, as in our approach. To explore RQ1, we use the \textit{all-features} upper bound that has access to both the human and machine features, and the \textit{machine-only} that uses no human features. To explore RQ2, we use a \textit{feature-selection} baseline using a greedy forward selection strategy (see \cite{tang2014feature} for an overview). To explore RQ3, we have 2 simple baseline algorithms: \textit{plausible-classes}, based on choosing features that differentiate between the most likely classes, and \textit{surprising-features}, based on uncertainty in the predicted values of the human features. See Appendix for more details.

\begin{figure}
 \centering
 \includegraphics[width=0.49\linewidth]{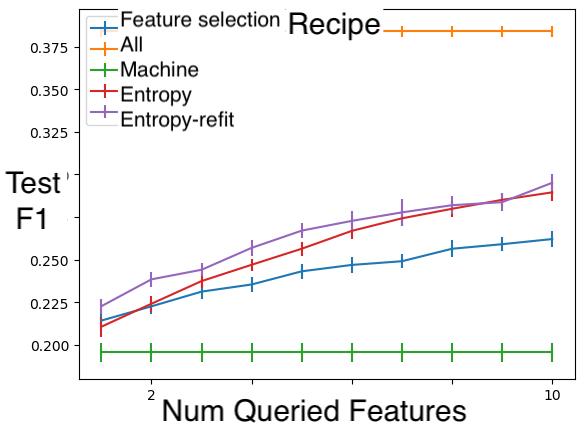}
 \includegraphics[width=0.49\linewidth]{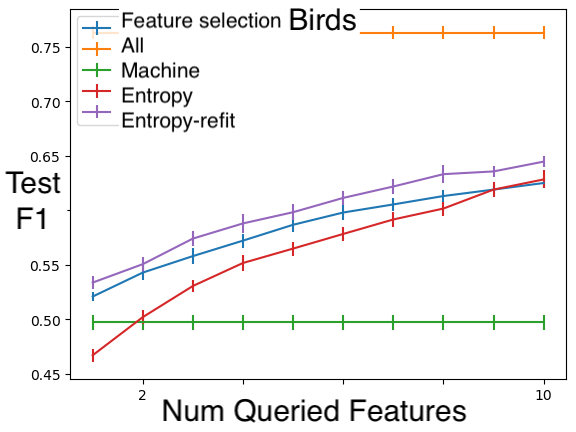}
 \caption{Test f1-score as a function of $B$ for both instance-wise feature selection methods, baselines and upper bound in \textit{recipe} dataset (left) and \textit{birds} dataset (right). Error bars are standard errors over 10 random restarts. \textit{all-features} performs best and \textit{machine-only} worst with the methods using a subset of human features in between. \textit{entropy-retrain} outperforms \textit{feature-selection} in both domains.}
\label{fig:recipe-results}
\end{figure}

\subsection{RQ1 Results}

\paragraph{The full set of features outperforms \textit{machine-only}, suggesting that the human features contribute substantially to predictive performance.} Figure~\ref{fig:recipe-results} shows the mean and standard deviation of the test f1-score for both methods computed over 10 random restarts of the train/valid/test splits and any subsampling. This is reported as a function of the number of features queried. In both datasets, \textit{all-features} performs substantially better than \textit{machine-only}. In the recipe dataset, the gap in f1-scores between these methods is 0.18, while in the birds dataset, the gap is 0.26. This suggests that there is substantial improvement to be gained from finding ways to use the human features in predictions.

\paragraph{\textit{entropy-retrain} is able close a substantial portion of the gap between \textit{all-features} and \textit{machine-only} within just 10 queries.} In both datasets, the \textit{entropy-retrain} method substantially improves performance over \textit{machine-only}, closing half or more of the performance gap by the 10th query in both datasets. This is true despite the 10 queries covering only 12\% of the human features in the \textit{recipe} dataset and 8\% of the features in the \textit{birds} dataset. This suggests that querying a small number of relevant human features is a viable strategy to substantially improve performance of ML models that have access to only a subset of relevant features. 

\textbf{Takeaways} These results suggest that in both of these datasets, a small number of the held-out features substantially improve predictive performance over the \textit{machine-only} baseline. This indicates that exploring effective ways to incorporate human features into predictions is worthwhile, and provides evidence towards the utility of approaches that query a small number of human features to improve predictions.

\subsection{RQ2 Results}

\paragraph{Our proposed \textit{entropy-retrain} approach outperforms \textit{feature-selection} in both datasets after sufficient queries, and never underperforms it.} In the \textit{recipe} dataset, \textit{entropy-retrain} outperforms \textit{feature-selection} starting from the 2nd query onwards. In the \textit{birds} dataset, \textit{entropy-retrain} performs comparably to \textit{feature-selection} for the first 6 queries, then outperforms it for the last 4. The performance improvement is particularly marked in the recipe domain, where \textit{entropy-retrain}'s f1-score after 10 queries is 0.3 compared to only 0.26 for \textit{feature-selection}. This suggests that querying unique human features for each instance can improve performance over a shared set of features queried for the entire dataset. It also suggests that our proposed retraining heuristic in \textit{entropy-retrain} is important for achieving these performance gains.

\begin{figure}
 \centering
 \includegraphics[width=0.5\linewidth]{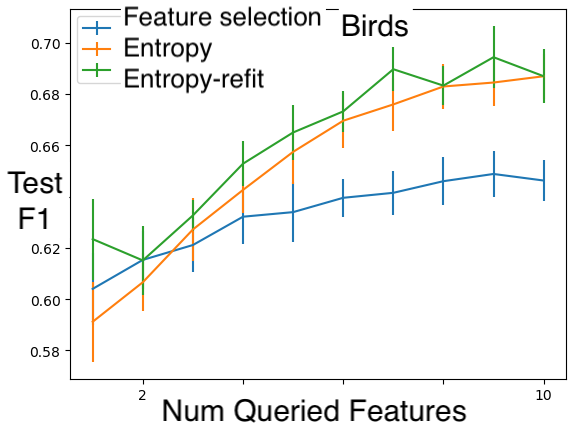}
 \captionof{figure}{Test f1-score on \textit{birds} as a function of $B$ after adding first 6 \textit{feature-selection} queries to machine features and re-running. Error bars are standard errors from 5 random restarts. By query 6, both entropy methods substantially outperform \textit{feature-selection}.}
 \label{fig:birds-updated}
\end{figure}

\paragraph{In the \textit{birds} dataset, doing a round of \textit{feature-selection}, then running \textit{entropy-retrain} shows even more drastic improvements relative to \textit{feature-selection}; \textit{entropy-selection} shows similar but less marked improvements.} Figure~\ref{fig:birds-updated} shows the results of a follow-up experiment where we added the first 6 features selected by \textit{feature-selection} to the machine feature set, $X^m$, for the birds data, then re-ran \textit{feature-selection}, \textit{entropy-retrain} and \textit{entropy-selection} for 10 additional queries. These results show 5 random restarts. Here, we see that \textit{entropy-selection} and \textit{entropy-refit} perform similarly for the first 4-5 queries, then substantially outperform \textit{feature-selection} with an f1-score of 0.69 vs. 0.65. This suggests that for some datasets, individualized feature queries show a drastic improvement immediately, while for others, they may be more useful after querying an initial set of shared features through feature selection. Even in this case, \textit{entropy-refit} outperforms \textit{entropy-selection}. 

\textbf{Takeaway} These results demonstrate the importance of selecting distinct features for each instance, as this outperforms the global feature selection approach. This motivates developing more effective methods for instance-level feature integration to improve human-ML decision-making.

\subsection{RQ3 Results}

\begin{figure}
 \centering
 \includegraphics[width=0.49\linewidth]{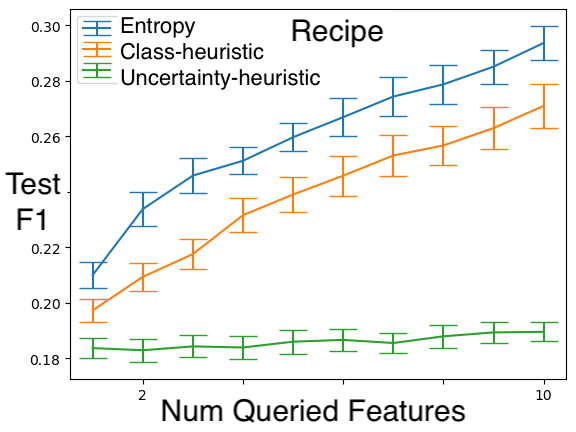}
 \includegraphics[width=0.49\linewidth]{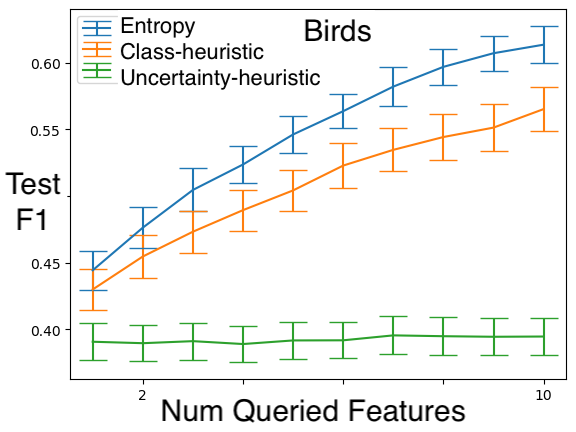}
 \caption{Test f1-score as a function of $B$ for both \textit{entropy-selection}, and the \textit{plausible-classes} and \textit{surprising-features} baselines in \textit{recipe} (left) and \textit{birds} (right) datasets. Error bars are standard errors over 10 random restarts. The \textit{entropy-selection} approach substantially outperforms the two baselines for choosing $q$, demonstrating that the additional complexity provides additional predictive value.}
 \label{fig:heuristic-results}
\end{figure}

\paragraph{In both datasets, the \textit{entropy-selection} approach substantially outperforms the \textit{plausible-classes} and \textit{surprising-features} baselines, suggesting that these are not sufficient to reach the good predictive performance of our proposed problem formulation.} Figure~\ref{fig:heuristic-results} shows the performance of the \textit{plausible-classes} and \textit{surprising-features} baselines compared to \textit{entropy-selection}. In both of these cases, the entropy approach substantially outperforms these baselines. This suggests that it provides additional utility compared to these simple algorithms for choosing $q$.

\textbf{Takeaway} These simple baseline algorithms for choosing the human features to query do not achieve the predictive performance of the \textit{entropy-selection} method used to solve our proposed problem formulation. This provides evidence that algorithmically querying the user for specific features may be more useful than allowing them to choose features to input.

\section{Conclusion}

Users of ML systems often have personal expertise about their lived experience that should influence predictions made about them, but existing approaches to incorporating that expertise only work if the user is a domain expert. We propose a new problem formulation called \textit{human feature integration} that only requires users to have personal expertise about the relevant human feature(s) rather than also requiring expertise about the prediction process. Under our problem formulation, the user can feed personal knowledge into the model, rather than feeding in their own prediction, or receiving an explanation of a prediction and trying to figure out what to do with it. Because of this, the user is not obligated to know \textit{how} their personal knowledge should be used in the prediction process in order to have it influence the prediction. \textbf{This problem formulation addresses a key technical gap of how a user with personal expertise but no significant domain knowledge about a prediction process can have their personal expertise considered in the output of an ML system.}

In this paper, we characterized the problem of human feature integration. Through a qualitative discussion and pedagogical example, we justified the importance of the problem and showed that it fills a gap left in the literature. Through a formal description of the problem and corresponding optimization objective, we laid out a framework for developing solutions for human feature integration, and key technical challenges that those solutions must address. Through an experimental evaluation of a simple operationalization of our framework in semi-realistic data, we demonstrated the feasibility of solving this problem; provided preliminary evidence of its effectiveness compared to alternative methods for incorporating personal knowledge in the form of features; and outlined a strategy for evaluation.

\section{Acknowledgements}

This work was funded by NSF GRFP grant no. DGE1745303; NSF IIS-2107391 and NSF IIS-1750358.

\bibliography{bibliography}
\bibliographystyle{named}

\appendix 

\section{Datasets} 

We use two real datasets in our experiments: a \textit{recipe} dataset
\footnote{\url{https://www.kaggle.com/datasets/kaggle/recipe-ingredients-dataset} train.json file} 
where the goal is to predict the cuisine of a recipe (e.g. Italian food) based on the ingredients, and a \textit{birds} dataset (Wah et al. 2011)
 where the goal is to predict the type of bird (e.g. crow) based on some crowdsourced attributes of an image of a bird. After pre-processing, the \textit{recipe} dataset consists of 6k instances (subsampled to that size for computational reasons); 120 features, and 20 classes, and the \textit{birds} dataset consists of 5k instances; 171 features; and 36 classes.

To facilitate running experiments with ground truth values of $X^h$, we split the feature sets between $X^m$ and $X^h$. We construct these feature splits so that $X^m$ consists of a smaller set of features that are ``simpler'' in some way, and $X^h$ consists of a larger set of features that are more ``complex''.  This is designed to simulate a setting where the human features are varied and add nuance to a core machine feature set that exists in the dataset.  In the \textit{recipe} dataset, we split the features so all single word ingredients (33 features) are in the machine feature set, and all 2$+$ word ingredients (87 features) are in the human feature set. For the \textit{birds} dataset, we split them so all non-color-related words (48 features) are in the machine features, and all color-related words (123 features) are in the human features. 

We preprocess the datasets as follows.  We preprocess the recipe dataset by removing instances without labels, then subsampling 15\% of the remaining instances, maintaining overall label distribution, for computational reasons. We then removed features with positive values for less than 100 instances. For the birds dataset, we use the coarse class labels, and remove instances with infrequent class labels, recorded for less than 50 instances. We then binarize features values at 0.5 (features are recorded as the fraction of agreement on MTurk responses about whether the feature is present in the image), and removed features with positive values for less than 100 instances. We split the data, class-balancing labels, into train/validation/test sets of sizes $\frac{2}{3}, \frac{1}{6}, \frac{1}{6}$ of the instances respectively.

\section{Method Implementation}

We describe 3 implementation details relating to our method.  1) How we solve Equation~\ref{eqn:loss}, 2) an implementation heuristic that we designed to improve performance, and 3) which functional forms we use to instantiate $f$ and $\hat{p}(X^h_d|X^m)$. 

\subsection{Solving Equation~\ref{eqn:loss}}
In practice, we solve the optimization sub-problem described in Equation~\ref{eqn:loss} by trying all possible values of $d$ that satisfy the constraint. This is a strategy linear in $D^h$, and is employed by many active learning approaches, including \cite{zhu2003combining}. Algorithm~\ref{alg:greedy-search} gives a full description of the procedure. 

\begin{algorithm}
\caption{This greedy search-based optimization procedure chooses, at each iteration, to query the human feature that minimizes the expected marginalized entropy given the feature is queried.}\label{alg:greedy-search}
\begin{algorithmic}
\STATE{Given: $x^m, x^h$}
\STATE{$q \gets \{0\}^{D^h}$}
\FOR{$\texttt{b} \in \{1,...,B\}$}
 \STATE{$d^* = \argmin_{d \in \{1,...,D^h|q_d=0\}} \mathbb{E}_{p(x^h_d|x^m)}[H(f^{marg}(x^m, x^h, q + \mathbbm{1}^{D^h}_{d})]$}
 \STATE{$q \gets q + \mathbbm{1}^{D^h}_{d^*}$}
\ENDFOR
\end{algorithmic}
\end{algorithm}

\subsection{Implementation heuristic: Re-training with Query Mask}
\label{sec:retrain}

We now introduce a novel retraining heuristic that is necessary for this approach to outperform dataset-level feature-selection in some of the real data cases we study.   

The core intuition for this heuristic is that the parameters of the prediction function, $\theta^m, \theta^h, \phi$ are trained to expect the full set of human features at test time.  But in practice we only have access to $B$ of them for each instance.  The vanilla entropy method described above addresses this by marginalizing out unqueried dimensions when predicting, however we find that we can improve performance by retraining the prediction function to make predictions based on the optimized $q$ for the training set.

To make predictions with this heuristic, we use a prediction function $f^{zero}$ that zeros out the features not in $q$.  We define the prediction function $f^{zero}$ as follows:
\begin{equation}
 f^{zero}(x^m, x^h, q) = \sigma((\bar{\theta}^m)^T x^m + (\bar{\theta}^h)^T ( x^h \odot q ) + \bar{\phi} )
\end{equation}
.  The re-trained parameters $\bar{\theta}^m$, $\bar{\theta}^h$, and $\bar{\phi}$ are fit using a standard approach (i.e. the same approach used to train $f$) on the transformed dataset: $(X^m, ( X^h \odot Q ), Y)$. Here, $Q$ is the matrix of feature query masks fit using the approach described in Section~\ref{sec:approach} on the training set.

In order to implement the method described in Section~\ref{sec:approach}, we need to define functional forms for $f$ and the probability distribution $p(X^h|X^m)$ used to marginalize out the un-queried human features.  We describe these below.  We then describe hyperparameters.

\subsection{Functional forms} We implement $f$ as a logistic regression, where we make predictions as follows:
\begin{equation}
 \hat{Y} = \sigma((\theta^m)^T X^m + (\theta^h)^T X^h + \phi)
\end{equation}
Where $\theta^m$ and $\theta^h$ are weight vectors in $\mathbb{R}^{D_m}$ and $\mathbb{R}^{D_h}$ respectively, and $\phi$ is an intercept term in $\mathbb{R}$.  We fit the model using the scikit learn package \cite{scikit-learn}.  

We approximate $p(X^h|X^m)$ from the data at train time, using an independence assumption between the different dimensions of $X^h$ for computational convenience. We model each dimension $d$ of $X^h$ with an independent logistic regression model (by assumption, the features in $X^h$ are binary).  I.e. for a dimension $d$, we define:
\begin{equation}
 \hat{p}(X^h_d|X^m) = \sigma( w_d^T X_m + w_d^0 ) 
\end{equation}
where $w_d$ is a weight vectors in $\mathbb{R}^{D_m}$, and $w_d^0$ is an intercept in $\mathbb{R}$.  Each human dimension $d \in D^h$ has a unique set of weights $w_d$.  We fit these logistic regression models using the scikit learn package \cite{scikit-learn}.  

\subsection{Baselines}

We compare the proposed solutions to four baselines and one oracle upper bound.  To explore RQ1, we have baselines with all human features and with no human features.  To explore RQ2, we have a feature-selection baseline, and to explore RQ3, we have 2 simple baseline algorithms to choose the query set $q$. 

The \textit{all-features} and \textit{machine-only} upper bound and baseline allow us to explore the impact of incorporating human features into predictions.  The \textit{all-features} upper bound consists of a model with the same functional form as $f$ (logistic regressions) trained using both the machine and human feature sets, i.e. $X^m$ and $X^h$.  This is what would be possible if, at test time, we could query the human for \emph{all} their features rather than a subset.  The \textit{machine-only} baseline consists of a logistic regression model trained with only the machine feature set, $X^m$. This is the performance of the model with no access to human features.

The \textit{feature-selection} baseline allows us to explore an alternative approach to instance-specific feature queries where all instances have the same additional human features.  The \textit{feature-selection} baseline consists of a standard feature selection approach where the same $B$ human features are queried for all instances, rather than a distinct subset of human features being queried for each instance.  We implement it using a greedy forward selection strategy (see \cite{tang2014feature} for an overview) where, for each of the $B$ features to add, we re-train the model adding each remaining human feature and choose the feature with the best validation performance (computed using f1-score--the metric used in our results).

The \textit{plausible-classes} and \textit{surprising-features} are simple approaches to choosing the query set $q$ that allow us to explore the added benefit of the approach detailed in Section~\ref{sec:approach}.  The \textit{plausible-classes} baseline works by selecting features that differentiate the most likely classes according to $X^m$.  We implement this by identifying plausible classes as those where $p(y_i|X_m) > 1/K$ where $K$ is the number of classes and $1/K$ is the uniform probability.  We then rank features by the largest magnitude of any weight associated with this feature across classes.  The \textit{surprising-features} baseline works by querying the dimensions of $X_h$ that cannot be accurately predicted given $X_m$.  Unlike the method described in Section~\ref{sec:approach}, this does not take into account how much these uncertain features may influence predictions.  We operationalize this baseline by querying features in the order of the predicted uncertainty in their value--$|0.5 - p(X_h^i|X_m)|$. 

\section{Experiment Hyperparameters} We use the following hyperparameters in our experiments.  We set $B=10$ as this is a manageable number of features for a single user to provide.  We set the number of Monte Carlo samples $S=5000$ to minimize this source of approximation error.  In the logistic regression models, we use a multinomial loss, the SAGA solver and we set the maximum number of iterations to 5000.  We perform a hyperparameter search over the following choices: penalties: [l1, l2 and none], inverse regularization strengths [ 0.01 , 0.1 , 1. , 10. , 100. ], and class weighting schemes [none, balanced].  We select hyperparameters to maximize f1-score (the metric we report) on the validation set, except in the case of the models for $p(X^h|X^m)$, where log-loss is minimized instead. This is to encourage those models to accurately model the probability distribution they approximate.

\section{Additional Results}

\paragraph{\textit{entropy-selection} has variable performance compared to \textit{feature-selection}, outperforming it in some cases and underperforming it in others.}  Figure~\ref{fig:recipe-results}, in the main text, additionally shows the performance of the vanilla \textit{entropy-selection} method introduced in \cite{quost2021decision} and the \textit{feature-selection} method.  In the recipe dataset, \textit{entropy-selection}, performs similarly to \textit{entropy-retrain}, which allows it to substantially improve over \textit{feature-selection} starting on the 4th query. In the birds dataset, \textit{entropy-selection} actually performs worse than \textit{feature-selection} for the first 6 queries, then performs similarly. This suggests that the vanilla method proposed in \cite{quost2021decision} without the retraining heuristic sometimes provides additional performance over dataset-level feature selection, but not always.  

\begin{figure}
\centering
\begin{minipage}{.48\textwidth}
 \centering
 \includegraphics[width=\linewidth]{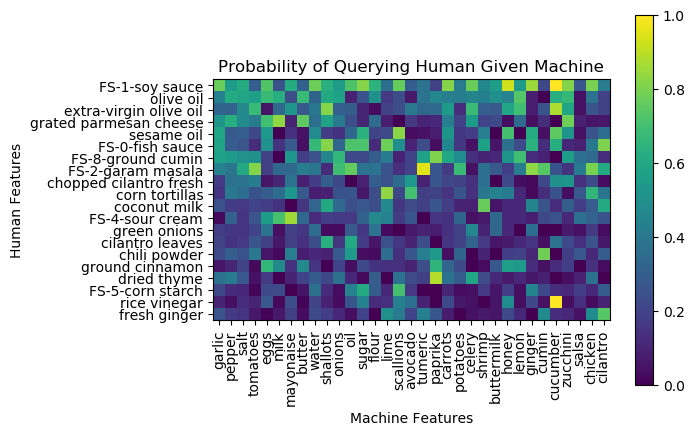}
 \captionof{figure}{Probability of querying a human feature (y-axis, top 20 sorted by \# times queried) given a machine feature (x-axis) in the instance in \textit{recipe}. Computed on test set for randomly chosen restart. ``cucumber''--``rice vinegar'' and ``turmeric''--``garam masala'' associations are sensible.}
 \label{fig:recipe-qualitative}
\end{minipage}
\end{figure}

\paragraph{Sensible correlations between machine features and human feature queries in the recipe domain suggest reasons for performance improvements from individualized feature queries.}  Figure~\ref{fig:recipe-qualitative} shows a heatmap of the probability a specific human feature (y-axis) will be queried using \textit{entropy-selection} given that a specific machine feature (x-axis) is observed for the instance (computed on the test set for 1 randomly chosen random restart). Strong relationships between sensible ingredient pairings include ``cucumber'' and ``rice vinegar'', which makes sense because the ingredients commonly co-occur in various east Asian cuisines but ``cucumber'' is more widely used, and ``turmeric'' and ``garam masala,'' which makes sense because again, ``turmeric'' can be used in various cuisines, but the 2 ingredients frequently co-occur in Indian cuisine. This suggests that feature queries made by \textit{entropy-selection} are informed by the machine features in sensible ways.  

\end{document}